\documentclass[runningheads]{llncs}
\usepackage[utf8]{inputenc}
\usepackage{graphicx}
\usepackage{caption}
\usepackage{textcomp}
\usepackage{subcaption}
\usepackage{array}
\newcolumntype{P}[1]{>{\centering\arraybackslash}p{#1}}
\usepackage{lipsum}

\title{Improving Welding Robotization via Operator Skill Identification, Modeling, and Human-Machine Collaboration: Experimental Protocol Implementation}
\titlerunning{Improving Welding Robotization via Operator Skill Identification}
\author{Antoine LENAT\inst{1,2}, Olivier CHEMINAT\inst{2}, Damien CHABLAT\inst{1}, Camilo CHARRON\inst{1,3}}
\authorrunning{A. LENAT et al.}
\institute{
$^1$Nantes Université, École Centrale Nantes, CNRS, LS2N, UMR 6004 \\
44000 Nantes, France\\
$^2$ Université Rennes 2, Rennes, France\\
$^3$ CETIM, 74 Rte de la Jonelière, 44300 Nantes \\
\email{\{Antoine.Lenat, Camilo.Charron, Damien.Chablat\}@ls2n.fr}}
\institute{
$^1$Nantes Université, École Centrale Nantes, CNRS, LS2N, UMR 6004 \\
44000 Nantes, France\\
$^2$ CETIM, 74 Rte de la Jonelière, 44300 Nantes \\
$^3$ Université Rennes 2, Rennes, France\\
\email{\{Antoine.Lenat, Camilo.Charron, Damien.Chablat\}@ls2n.fr}, Olivier.Cheminat@cetim.fr}

\date{January 2024}

\begin{document}

\maketitle

\begin{abstract}
The industry of the future, also known as Industry 5.0, aims to modernize production tools, digitize workshops, and cultivate the invaluable human capital within the company. Industry 5.0 can't be done without fostering a workforce that is not only technologically adept but also has enhanced skills and knowledge.
Specifically, collaborative robotics plays a key role in automating strenuous or repetitive tasks, enabling human cognitive functions to contribute to quality and innovation. 
In manual manufacturing, however, some of these tasks remain challenging to automate without sacrificing quality. In certain situations, these tasks require operators to dynamically organize their mental, perceptual, and gestural activities. In other words, skills that are not yet adequately explained and digitally modeled to allow a machine in an industrial context to reproduce them, even in an approximate manner. Some tasks in welding serve as a perfect example. Drawing from the knowledge of cognitive and developmental psychology, professional didactics, and collaborative robotics research, our work aims to find a way to digitally model manual manufacturing skills to enhance the automation of tasks that are still challenging to robotize. Using welding as an example, we seek to develop, test, and deploy a methodology transferable to other domains. The purpose of this article is to present the experimental setup used to achieve these objectives.
\end{abstract}

\section{How to analyze skills with psychology}
Skills are critical for problem resolution, this explains why literature about this subject is abundant, notably in psychology. We can find lots of definitions about skills, but they each share four mainstays that are well summarized in Vergnaud theory \cite[p.88]{vergnaud_theory_2009}. Skills are based on goals, rules to generate activity, operational invariants, and inference. When an operator, as a welder, is resolving a situation he is anticipating the evolution of the situation with and without his intervention, this is a goal. To interact with the situation, he does some action, gathers information, and controls, these rules generate an activity. He has to decide which information is pertinent and/or true, this is based on beliefs and implicit knowledge. These is operational invariants. All of this needs a computational activity from the brain to adapt to the situation, which we will name inferences. When some of these pillars are consistent from one situation to another, we obtain a scheme. Schemes are, in Vergnaud own words, ``\textit{the invariant organization of activity for a certain class of situations}``(p. 88). 
Hoarau \cite{hoarau_revealing_2018} gives examples of activity analyses made with scheme framework, where strategies used by operators are revealed by identifying consistent control loops from one situation to another. Moreover, the scheme can be adapted for specific fields like mathematics \cite{charron_conceptualization_2002}. 

The four elements of the scheme require consistent regulation to facilitate adaptation by the operator. Welders, for instance, must actively gather information on the current situation, enabling them to compare it with the predicted scenario. Any mismatch will trigger a need for immediate adaptive measures; cognitive or physical ones. We can observe that an operator evolves in a situation to resolve a problem, having a productive impact. At the same time, the ability to adapt to various situations while performing tasks contributes to a valuable experiential learning process, making constructive growth for the individual.  These regulations can operate in a closed loop for action or a longer loop in planning, for example.
Finally, a scheme can be a good way to understand skills, giving an opportunity to have a modus operandi transferable from one field to another. This modus operandi began by understanding the profession's analysis. In doing so, we can propose a set of variables and situations to analyze the four fields of the scheme. When we have measured variables in these situations, we can search for invariant organization that are schemes. These schemes are a good way to analyze and describe skills, abilities and competencies. Each of this terms describe one part of the schemes while the latter include dynamical organization and adaptation to a specific class of situations. This is why we will use the term of schemes in a way that incorporate skills, abilities and competencies.

\section{Welding Variables}
Studying welding revealed many parameters, around sixty, that can directly influence the bead or have an impact on the welder. We propose here a classification of these variables into the following categories: Human factors, Physico-chemical forces,  Process control parameters (electrical and shielding gas), Workpiece parameters, Manipulation parameters, Filler metal parameters, and Ergonomic parameters.

The choice of current, a critical parameter in welding, significantly influences welding energy, alongside welding speed and voltage. Manually selected by the welder, it becomes our first variable to consider. However, this setting is subject to variation based on the distance between the electrode (torch) and the workpiece to be assembled, which we refer to as stick out. Similarly, the torch angle impacts the distribution of this energy. Therefore, we need to have an estimation of the torch's position and orientation concerning the workpiece.

This orientation and position are directly influenced by the welder's choices and manipulation, notably via sick-out. Estimating how the latter is precisely carried out is crucial to understanding the human-specific aspect. That's why we need to focus on the complete body movement of the welder to estimate the welding speed and the variations that may be imposed. Moreover, good welding can be obtained with several combinations of parameters since they synergize. 

A major difference exists between the two welding processes we intend to study (GMAW and GTAW): the filler material deposition. In the first one, the filler metal is supplied by a melting electrode. Therefore, deposition quantity is directly linked to the welding current, nature of the filler material, and wire diameter. In GTAW, the welder manually introduces the filler metal with a filler rod: hence, they must manage the feed rate, deposition quantity, and the location of deposition (in the weld pool, in the arc, etc.).

Other choices are left to welders, although these choices are guided by standards regarding the dimensions of the parts and other constants. We particularly consider the choice of the number of passes, stops (stopping the bead and then restarting), length of filler metal exiting the GMAW nozzle before welding, and the precision applied to the preparation of parts like chamfering and grinding.

The human factor is thus crucial: an analysis of it seems relevant. We can explore the degree of investment, physical or cognitive fatigue, adaptation to discomfort related to posture or welding fumes, eye position, and reaction time.

\begin{figure}
    \centering
    \includegraphics[width=\linewidth]{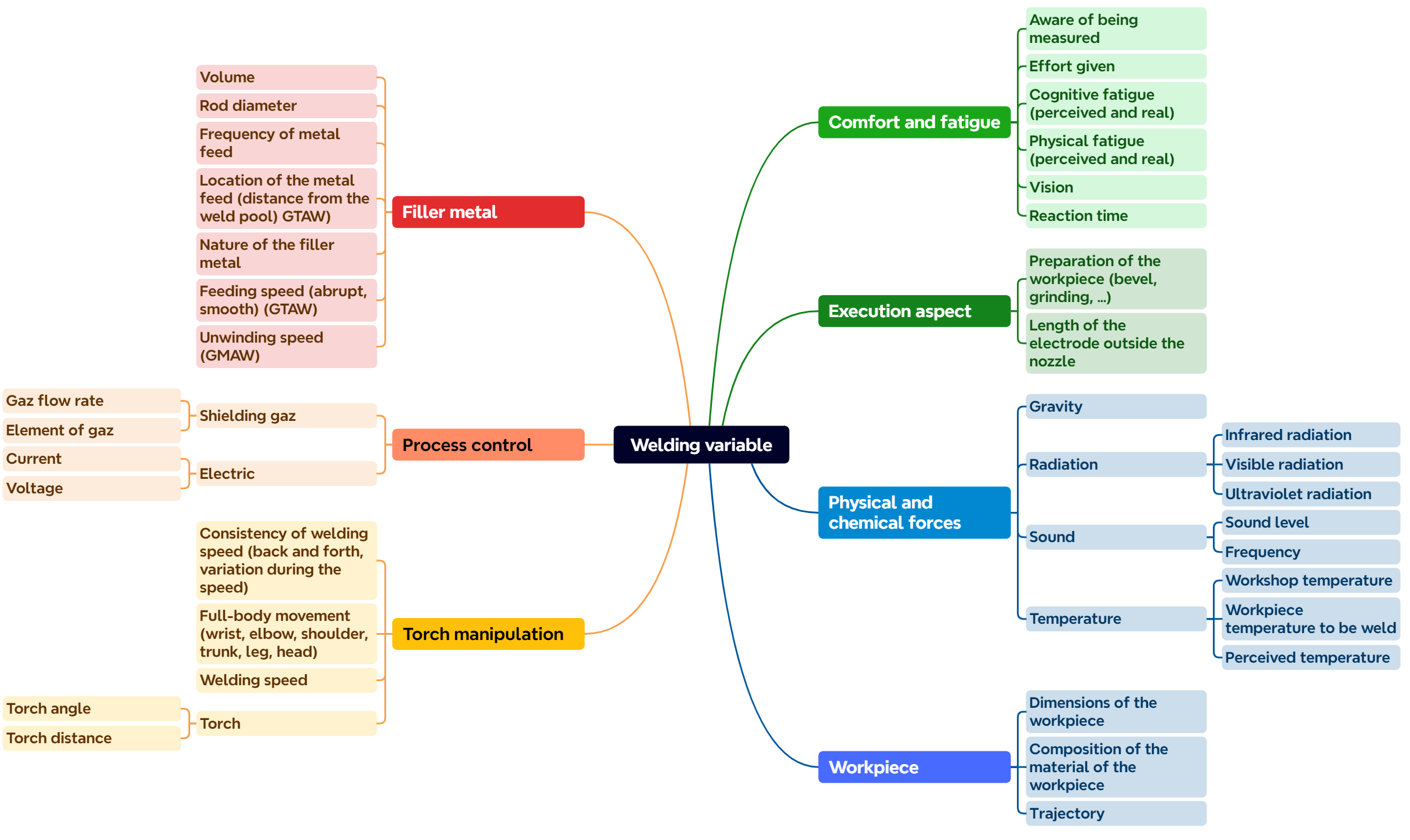}
    \caption{Mind map of welding parameters to study}
    \label{fig:MindMap Variables étudiées}
\end{figure}

The piece to be assembled will govern the welding parameters; therefore, the dimensions and nature of the piece materials must be specified. Among these parameters, we can also look into shielding gas, especially the nature of the gas and its flow rate. The welding to be performed will depend on the position of the two pieces to be assembled. If welding vertically, we must consider the variation of forces at play, including conduction in the metal pieces and gravity on the weld pool. All these variables are shown in figure \ref{fig:MindMap Variables étudiées}.
 
The next section will introduce the measurement tools that we are considering for the study of these parameters.

\section{Measuring instruments}
After defining relevant parameters, we must choose measurements guided by the question: how can a welder achieve a weld that a robot cannot?

Among the various measurement systems, we find those associated with electrical parameters. These parameters govern the supply of energy necessary for melting the edges and creating the weld:
(i) Welding generator display, (ii) Voltage (as a measure of the torch/workpiece distance), and (iii) Multimeter (for current and voltage)

Welding relies on the supply of energy and, in some cases, on the supply of metal. To analyze the volume of metal supplied in GTAW, we plan to graduate the metal rods to have reference points when filming. The dimensions of the GTAW filler rod must also be documented, especially its diameter. The analysis of the filler metal can thus be carried out using:
(i) Welding camera (Cavitar\textsuperscript{\textcopyright} \footnote{https://www.cavitar.com/applications/welding-imaging/}), (ii) Measuring wire-feed roller, and (iii) Ruler.

Next, we have a set of measurements related to the workpieces to be assembled; the dimensions of these are taken into account to define the parameters. Additionally, we can verify pre-welding modifications (chamfering, grinding), and finishing touches:
(i) Surface condition (roughness), (ii) Angle protractor or Goniometer, and (iii) Ruler/caliper.

Measurements of the human body can be used for two purposes: Tracking movement and analyzing physiological functions :
(i) Inertial sensors, (ii) Stopwatch (for welding speed), (iii) EEG (Electroencephalogram), (iv) ECG (Electrocardiography, (v) EMG (Electromyography), (vi) Eye-tracking, (vii) Optical motion capture, and (viii) VO2 (Oxygen consumption).

Finally, we can analyze the weld pool during welding to obtain, notably, its dimensions (width and/or thickness). Sound analysis of welding is also relevant:
(i) Welding camera (Cavitar \textsuperscript{\textcopyright}), (ii) Laser, (iii) Microphone / acoustic sensor, and (i) Ultrasound.

Video capture can be valuable for several reasons, particularly to easily visualize information in a qualitative manner. This allows us to analyze the addition of filler metal in GTAW \cite{manorathna_human_2017}, observe variations in torch manipulation, identify body segments involved during welding, estimate arc length, and more.

To complete these measurements, a survey or interview can be developed to understand the welder's choices and gain insights into how they justify their actions and decisions.
\section{Analysis of Measurement Tools}
We do not intend to delve into an analysis involving machine learning or neural networks. These techniques require extensive training and, thus, a large database that we do not possess. Conversely, we prefer to focus on a detailed understanding of a small sample and describe their skills within the schemes framework.

\subsection{Electrical Parameters: Process Control}
Arc welding is governed by its electrical parameters, particularly current and voltage. These two parameters, combined with welding speed, allow the calculation of welding energy, an interesting indicator to understand the process evolution.

The set current is directly adjusted by the operator in the generator; however, this value fluctuates with the distance between the welding torch and the workpiece. Some welding machines offer the capability to measure this physical quantity and store it in memory, enabling precise post-analysis. This measurement can be made with a multimeter, Hall effect \cite{manorathna_human_2017} or WeldQAS.

Voltage is in synergy with current and also varies depending on the distance between the torch and the workpiece (stick-out). Thus, this distance can be used as a control parameter on voltage by the welder during the process. Multimeter and WeldQAS can also be used to measure these quantities, especially voltage \cite{manorathna_human_2017}.
\subsection{Filler Metal: A Critical Process Difference}
For GMAW welding, using a consumable electrode, the filler metal's direct dependence on current ensures consistent wire feed rate. Analyzing it involves using a measuring roller integrated with the generator's unwinding rollers to derive the feed rate. Coupled with wire diameter, this yields feed volume. However, data gaps persist, notably the wire feed angle, obtainable through motion capture leveraging IMUs (Inertial Measurement Unit). Zhang \cite{zhang_measurement_2014} used IMUs to measure the inclination of GTAW's torch, but the same idea can be used in GMAW for torch and filler metal.

In GTAW, the welder manually holds the filler rod, rendering measuring roller use impractical. Some motion capture methods are disqualified as they disrupt the filler rod's weight and the welder, presenting challenges for analysis. However, nearby cameras can effectively capture filler metal angle and frequency. Although the uniform filler rod makes added metal volume inconspicuous, graduation is considered for visual assistance in video, aiding feed volume calculation.

We will use a welding camera to measure these angles based on \cite{manorathna_human_2017} and more parameters such as frequency or transfer mode \cite{goncalves_e_silva_effect_2019,silwal_effect_2018}.
\subsection{Sound: A Characteristic Emission}
The welding process produces noise during the operation, dependent on numerous parameters, both in the audible range and inaudible acoustic emissions, as reported by \cite{wang_analysis_2009}. Some welders seem to use this sound as an indicator \cite{manorathna_human_2017,saini_investigation_1998,wang_feature_2011}.
Although the sound may vary significantly based on the nature and thickness of the workpieces to be welded, the welder may adapt and use it as an indicator to regulate the process. \cite{wang_analysis_2009}, particularly caution on the microphone angle and gas flow. Therefore, we consider the study of sound relevant and will use a microphone during our experiments. The parameters studied will include the frequencies and amplitudes of the sound signal; frequency domain analysis using a Fourier transform is also relevant, as shown by \cite{saini_investigation_1998,wang_analysis_2009}.

Some people propose estimating the depth (penetration) of the weld pool based on sound, as demonstrated by \cite{pal_monitoring_2011}. Sound appears to be an interesting possibility for real-time control of the welding process \cite{lv_investigation_2011}. Sound offers valuable characteristics for studying the welder's schemes, dimensions of the weld pool, and has a high potential for automating certain tasks currently not feasible for robots. Therefore, sound analysis is highly recommended.
\subsection{Temperature: Anticipating Cooling}
The workpiece temperature, coupled with the cooling time, governs the metallurgical structure of the weld bead; these details can be found in phase diagrams. These diagrams predict the phases obtained after thermodynamic equilibrium based on temperature and the chemical composition of the alloy. Temperature is a critical parameter depending on the welding energy supplied, material nature, and the environment.

The measurement of temperature in welding is described in the standard ISO 13916 \cite{iso_13916_soudage_2018}, which mandates measuring the temperature at a specific distance from the weld bead. Various measurement tools are available with different levels of precision, including:
(i) Thermosensitive products (e.g., pencils or paints) (TS), (ii) Contact thermometer (CT), (iii) Thermocouple (TE), and (iv) Optical or electrical devices for non-contact measurements (TB).

In our case, the electric arc in the GTAW process reaches 3000 °C in a room temperature environment; metals will rise to around 300 °C. With such a wide measurement range, we cannot expect high precision. We can still measure the temperature of the bead for some information with optical measurement. Temperature can, indeed, be a way for an operator to analyze the welding process (directly or indirectly).
\section{Weld pool Dimensions: A Visible Factor for the Welder}
The weld pool represents the liquid part of the workpieces being assembled, with a very high light intensity; it is the most visible element, perhaps the only one, that the welder sees during the process. Its dimensions directly influence the shape of the weld bead since it's forming during its cooling.

Dimensions include the visible surface and the penetration depth. Applying too much energy in one area risks melting the piece throughout its thickness and creating a hole. It can be an interesting element to measure. However, the strong contrasts imposed by the weld pool and ambient lighting make video measurements challenging. We will explore different approaches.
\subsection{Laser Measurements}
Laser techniques analyze light diffraction to measure the surface dimensions of a weld pool \cite{alvarez_bestard_measurement_2018,kovacevic_sensing_1996}. However, articles often neglect weld pool penetration analysis. Ensuring data continuity, particularly in-depth, is challenging; analyzing weld bead thickness is feasible, but penetration rate assessment remains elusive. The technique lacks portability and exhibits variable performance. This method is expensive, complex to implement, and appears irrelevant to our study, considering its drawbacks.
\subsection{Ultrasonic Measurements}
Ultrasonics can be used to determine weld pool dimensions, especially its depth \cite{hardt_ultrasonic_1984,alvarez_bestard_measurement_2018} mention several articles using this technique, particularly to obtain weld pool depth. However, the heating of the workpiece and sensors limits the use for real-time (in-line) process control. Ultrasonics are effective in determining boundaries between states of matter in a welding process. Ultrasonics can also be coupled with a laser system described earlier \cite{mi_real-time_2006}, or with data such as intensity, voltage, or sound. Finally, studies on weld beads using ultrasonic techniques are generally proposed to detect common defects.

\subsection{Camera Measurements}
Cameras are very sensitive to strong contrasts, as mentioned; however, it is interesting to note that image processing can be used to obtain weld pool images. We mentioned Weez-U Welding\textsuperscript{\textcopyright}, which use a camera mounted on the robot to capture multiple successive images with different intensity filters before compiling them. Some authors use high-speed cameras with significant magnification to observe weld pool solidification \cite{chiocca_etude_2016,delapp_quantitative_2005}.
Some cameras offer direct filtering to observe through the arc radiation; this is notably the case with the Cavitar \textsuperscript{\textcopyright}, which can help measure the dimensions of the weld pool. Since we can use Cavitar to measure the angle of the filler rod and torch with the workpiece, we can use it to measure the dimensions of the weld pool.  
\section{Operator Motion Capture}
We aim to compare two elements: what the welder precisely executes and what justifies their actions. This involves studying the welder's movements to understand the welding speed, selected intensity, regularity of the motion, etc. These details are then compared with the reasons behind the actions, exploring why a variation in movement occurs at a specific moment and what information is perceived. Analyzing the operator's beliefs and knowledge is crucial, considering both implicit and explicit dimensions.

Among the various measurement instruments, we have particularly focused on motion capture. The welder's movements reflect decisions, personality, and the ability to interact with the environment to adapt to a situation. In our case, the situation involves performing an electric arc weld, a specialized process where the quality of the result is not directly observable. Moreover, the electric arc produces intense visible radiation and electromagnetic disturbances. It is essential to keep this in mind when analyzing measurement tools.

To capture human body movement, we need the orientation and position of different body segments to estimate joint angles. Tools considered for this purpose include:

\begin{itemize}
    \item Optical motion capture: the reflection of infrared light (wavelength greater than 800 nm) on passive markers to calculate the spatial position of markers. This system ensures high precision.
    \item IMU: a system consisting of an accelerometer and a gyroscope to measure acceleration and angular velocity, respectively. These data are then used to calculate the spatial position and orientation of a segment to which the IMU is attached. They are known for their high acquisition frequency and reasonable cost.
    \item RGB video: standard color video (Red Green Blue) that can be used to define the outlines of a subject. It can be coupled with an infrared camera to provide more information on the distance between a subject and the camera; this is then referred to as a depth camera. It combines ease of implementation and reasonable cost.
    \item Depth camera: estimation of the distance to an object using stereographic vision (double camera). Some depth cameras use infrared radiation, similar to optical motion capture. It allows reproducing movement in a three-dimensional environment.
\end{itemize}
\subsection{Optical Motion Capture}
Motion capture involves a set of cameras (transmitter/receiver) emitting light in near-infrared (IR) (wavelength 820 nm). The arc welding radiation spectrum is primarily in the visible range and extends into the UV. However, there are emissions in the IR range, with a peak observed at 817 nm, which may introduce noise into our tools. More information on radiation is available in articles by \cite{gourzoulidis_photobiological_2023,thomas_d_tenkate_optical_1998}, \cite{weglowski_investigation_2007}.

Optical motion capture is currently used in welding with virtual or augmented reality environments \cite{mueller_intuitive_2019,ong_augmented_2020}. The limitation we can highlight is the virtual aspect of welding, relying on models that may not be as precise as reality. An article by \cite{erden_identifying_2009} proposes performing this measurement by creating occlusion between the cameras and the arc using an opaque box based on measurements made by \cite{van_essen_identifying_2008}. However, these measurements suffer from data loss and noise due to radiation.

We conducted experiments to observe the impact of the welding process on optical motion capture measurements with a SmartTrack 3/M from ART. This tool is a portable motion capture camera, easily transportable in a workshop. The major drawback of this tool is its limit to 4 markers.

We tried to estimate the noise made by the electric arc on motion and IMU. Therefore, we decided to place two markers as fixed references on a linear support on the table. These two markers will be named body \#0 and \#1. Another one was used for measure and get placed on the Weez-U\textsuperscript{\textcopyright} robot arm. This one will be named body \#2. The experiment involved linear motion with torch ignition at the midpoint and a constant speed. The aim was to confirm data linearity by monitoring any mid-displacement variations, ensuring no interference from the electric arc. Moreover, our fixed reference (body \#0 and body \#1) should be measured during the whole welding process.

\begin{figure}[!ht]
    \centering
    \includegraphics[width=0.7\linewidth]{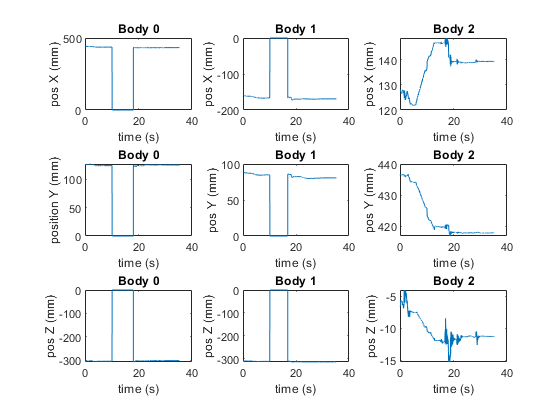}
    \caption{Welding arc noise test measurements on optical motion capture}
    \label{fig:Mesures_Bruit_IMU}
\end{figure}

Figure ~\ref{fig:Mesures_Bruit_IMU} illustrates the position according to a Cartesian coordinate system located at the center of the camera for the three present references. The trials measure the robot's positioning throughout the trajectory, unaffected by torch ignition. However, for the two fixed bodies, we observe a loss of vision (characterized by the return of positions to 0). This effect is not systematic during the trials.

In conclusion, these trials indicate good measurement stability after arc ignition if detection is ensured. However, some markers may experience occlusion if close to the axis between the camera and the electric arc, with the arc's radiation hindering marker detection. The solution appears viable, but careful attention should be paid to the camera's placement relative to the welder.

A new trial was conducted to measure the welder's arm movements. We placed a marker on the forearm and another on the same arm of the operator. We calibrated a new sensor, placing it out of the field of vision to check for possible spurious detection.
\begin{figure}
    \centering
    \includegraphics[width=0.7\linewidth]{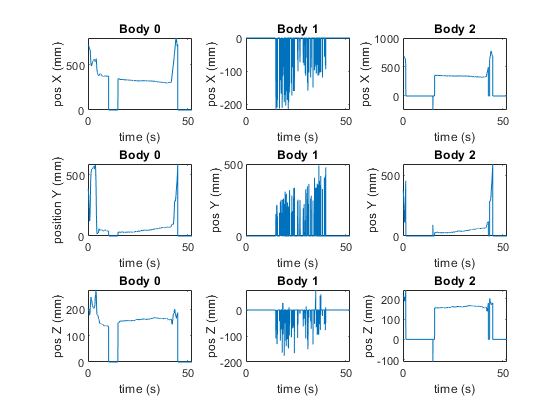}
    \caption{Results of capturing the movement of a welding arm}
    \label{fig:MoCap Bras Soudeur}
\end{figure}

The results are available in Figure \ref{fig:MoCap Bras Soudeur}. The absent calibrated body is body \#1; we observe its detection due to radiation produced by the electric arc. This emphasizes the need to pay attention to the camera's positioning relative to the welder and the workpiece.

The other two bodies (body \#0 and body \#2) are generally well-detected, but body \#0 is temporarily undetected (approximately 5 seconds). Thus, the arc introduces a significant source of noise.

Combined with the limit of four markers, this solution, despite its considerable advantages in precision and portability, seems incongruous with our objectives. However, potential avenues can be explored:
\begin{enumerate}
\item To address the low sensor count, acquiring an additional SmartTrack camera and synchronizing data acquisition could be considered.
\item To mitigate the disturbance from the electric arc, placing a band-pass filter in front of the lenses is an option. However, this may introduce optical path drift and data blurring. Additionally, as the arc and camera radiation are close, the band-pass filter doesn't guarantee satisfactory results.
\end{enumerate}
\subsection{Inertial Measurement Units (IMUs)}
Inertial Measurement Units employ a gyroscope and an accelerometer to provide angular velocity and acceleration of a body. This theoretically enables obtaining a segment's position through double integration of accelerometer data and its orientation through gyroscope data integration. However, the high sensibility of IMU makes acceleration data subject to noise. IMU are also used to obtain angle by measuring the acceleration of gravitation and coupling with data from the gyroscope by signal processing. 

Some researchers utilize these measurement tools in welding, including \cite{zhang_measurement_2014,zhang_welding_2017}. To assess IMU immunity to the measurement environment, we coupled the SmartTrack experiment with an inertial sensor attached to the Weez-U\textsuperscript{\textcopyright} arm. The sensor is positioned on a segment further from the torch for ease of implementation using an existing strap.

Similar to the initial observation with optical motion capture, no variations are observed during arc ignition. Although the trajectory is coded for constant speed, we note jerks attributed to the robot's architecture.

\begin{figure}
    \centering
    \includegraphics{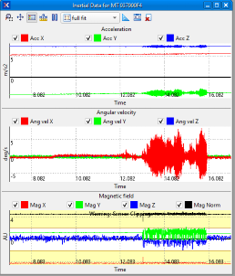}
    \caption{Inertial measurements with acceleration, angular velocity, and magnetic field according to a Cartesian reference.}
    \label{fig:Bruit IMU}
\end{figure}

On figure \ref{fig:Bruit IMU} we observe significant angular velocity and acceleration noise when the torch ignites, initially attributed to the magnetic field, but now believed to be the robot's vibrations during welding.

A follow-up experiment with the IMU on both a table and the floor aimed to confirm this hypothesis. The goal was to check if displacement was measured while the sensor was stationary. On the table, during arc ignition (identified by the magnetometer), the accelerometer displayed a highly noisy signal. On the floor, these signals were absent, demonstrating the sensor's immunity to disturbances from the electric arc.

Inertial Measurement Units appear promising for capturing human body movement. For a complete body, 15 sensors are necessary (head, neck, lumbar, two arms, two forearms, two hands, two thighs, two shins, two feet). Manufacturers offer suits providing joint data directly from these measurements, such as Movella\textsuperscript{\textcopyright} (with Xsens IMUs) or Teslasuit\textsuperscript{\textcopyright}, the latter potentially suitable for comprehensive testing.
\subsection{RBG video}
Traditional camera video is qualitatively valuable and can be used quantitatively by outlining the operator's contour for posture analysis, especially with a depth camera. However, this application is usually used in classification more than precise measurement of articulation's angle. Moreover, the intense light interference from the electric arc complicates this task. Cameras can provide additional insights into filler metal management or weld pool dimensions. Despite challenges, some industries use videos with extensive processing, capturing successive images with varying intensity filters and combining them to retain relevant portions, reducing significant contrast. Video capture remains beneficial for qualitative aspects, facilitating easy action visualization and data processing validation through comparisons with obtained data graphics. They can be used to take all the working scenes or be placed on the operator to have his field of view.
\subsection{Depth camera}
\begin{figure}[!ht]
    \centering
    \begin{subfigure}{0.4\textwidth}
        \includegraphics[width=\textwidth]{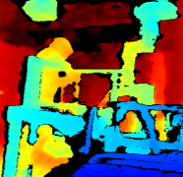}
    \end{subfigure}
    \begin{subfigure}{0.4\textwidth}
        \includegraphics[width=\textwidth]{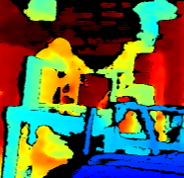}
    \end{subfigure}
    \caption{Depth camera scan difference with (left) and without (right) electric arc}
    \label{fig: DepthCamera}
\end{figure}

The depth camera was explored to capture an operator's contours and movements. This involves substantial data processing and multiple cameras to estimate the three-dimensional positions of all segments, along with the risk of occlusions. Tests were conducted with an Intel RealSense D400 camera using stereoscopy.
The two images in Figure \ref{fig: DepthCamera} show depth capture without an electric arc on the left and with an electric arc on the right. While the quality is good, distinguishing the workpiece, the background robot, the operator, and the torch, the image significantly distorts when an electric arc is lit. Despite its interest, this technology lacks precision and is affected by electric arc noise, rendering it unsuitable for our study.

\subsection{Measurement by force sensors/pressure plate}
A force plate is a tool in the form of a thin platform placed on the ground. It provides information on the distribution of forces and moments applied to support surfaces, offering an estimate of the subject's center of gravity changes while moving on it. Mainly used in gait analysis, these plates show the temporal evolution of vertical forces, providing insights into gesture deviation, direction, and balance. While redundant if we have access to the welder's complete motion, these plates can offer valuable information about the welder's balance and positional changes during welding but the welder needs to get on it during the process.
\subsection{Summary of Motion Capture Measurements}
Motion capture is dominated by optical techniques with passive markers. While highly precise, this technology requires large spaces with numerous cameras. The arc's radiation interferes with measurement wavelengths, introducing noise and occluding markers despite precautions \cite{erden_identifying_2009,van_essen_identifying_2008}. We utilized a portable SmartTrack 3 motion capture; despite its precision, the limitation to four markers is a significant drawback. If access to a complete room becomes available, we may reconsider this solution; currently, it is not feasible. Inertial sensors are cost-effective and offer high acquisition frequency and sensitivity. Though susceptible to noise, they provide precise data, making them relevant for our study \cite{ahmad_reviews_2013,gu_imu-based_2023}. Additional force plates can complement the setup. These techniques can also yield information on the torch's orientation and position.
\section{Physiological measurements: The body’s response}
Physiological measurements encompass signals from the body's electrical activity (muscles – EMG, heart – ECG, brain – EEG), eye movements (gaze tracking), oxygen consumption, emotional responses, etc. While some are valuable for user interface development, in our context, most won't provide significant information.
\subsection{Measurements of physiological electrical signals}
The analysis of muscular activity helps determine the muscles used in a gesture, providing a more detailed analysis of the movement and revealing associated muscle fatigue. Brain activity reveals cognitive load and the regions involved during an action, highlighting the preferential processing of one piece of information over another (approximately). Heart activity provides information on a person's stress; being also linked to physical activity, this complex data requires correlation with other information. Although it offers relevant details, it is not essential.

These three activities can be measured using the electrical signals produced by the human body:
(i) EEG: Electroencephalogram for the brain, (ii) ECG: Electrocardiogram for the heart, and (iii) EMG: Electromyogram for the muscles.

However, the generated intensities are low and highly influenced by the sensor's proximity to the skin. Additionally, electromagnetic disturbances from the welding arc can lead to significant interference. Due to the limited information provided by physiological measurements, their sensitivity to noise, and the substantial data processing required, we believe their use is not necessary.
\subsection{Gaze tracking}
The gaze tracking system, utilizing infrared oculometry, recalibrates the gaze direction on a video obtained from a camera located on the nose's edge. Some glasses supplement these data with a gyroscope to provide head orientation. Gaze tracking can offer valuable insights into an operator's behavior, revealing visual cues in the environment and providing information on cognitive information processing through fixation duration and gaze scanning rate. In our case, we aim to determine the gaze position concerning the weld pool's location, understanding whether the operator looks before or after the pool, fixates on it, or maintains a constant scan. However, the protective hood against the welding arc's light poses a challenge to using this technology, limiting its application in our study. Despite having oculometric data, we lack a detailed image, only obtaining a luminous point surrounded by darkness.

An attempt was made using Tobii Proglasses 2, provided by the University of Nantes. The camera quality remains good until the welding hood is worn. At that moment, the video becomes less usable and completely unworkable when the electric arc is initiated. Several possibilities could enable the use of this technique:
\begin{enumerate}
\item Modify the protective glass: Creating a protective glass with two different filter intensities, a weaker filter in front of the camera and a standard one for the eyes.
\item Utilize the gyroscope: Before initiating the arc, the trajectory can be identified based on the gap between the parts to be welded. Head tilt variations can be used to align the trajectory, considering the weld pool at the center of this line.
\item Move the camera: Placing the camera on the other side of the filtering glass may provide a clearer image; additional video processing might make the video usable.
\end{enumerate}
The first hypothesis poses safety risks if not executed correctly, and video quality might not guarantee to observe part positions. The second hypothesis is sensitive to gyroscope data drift. The third hypothesis involves challenges such as the difference in the field of view between the camera and eyes, along with data processing difficulties.

Eye tracking is used in welding, especially in conjunction with virtual reality. Despite the potential benefits of eye-tracking measurements, ensuring effective utilization of this measurement tool cannot be guaranteed.
\section{Summary and conclusion of the measurement tools}
In summary, considering all the measurement tools mentioned here and their potential use for studying the welder's schemes, we propose a matrix of possible utilization (see Table \ref{tab:measurement}). This matrix indicates whether a measurement tool is applicable (yes or no) in our case. If a tool is considered feasible under certain conditions, we may note "perhaps" with an explanation.

\begin{table}[ht!] 
    \centering
    \begin{tabular}{|P{2.3cm}|P{1,9cm}|P{1,9cm}|P{1,8cm}|P{3,6cm}|}
    \hline
        Measurement Tool & Applicable in our study & Under  conditions & Not applicable & Comments \\ \hline
    \hline
        Welding display & X &  &  & \\ \hline
        Cavitar\textsuperscript{\textcopyright} &  X &  &  & \\ \hline
        Non-contact thermometer & X &  &  & \\ \hline 
        ECG &  &  & X & \\ \hline 
        EEG &  &  & X & \\ \hline
        EMG &  &  & X & \\ \hline
        Eye tracking &  & X &  & Maybe with adaptation of the protective glass\\ \hline
        Measuring roller & X &  &  & Applicable in GMAW, not in GTAW\\ \hline
        IMU & X &  &  & \\ \hline
        Laser &  & X &  & Expensive \\ \hline
        Micro/acoustic sensor  & X &  &  & Noise and risk of occlusion \\ \hline
        Motion capture &  & X &  & \\ \hline
        Multimeter & X &  &  & \\ \hline
        Oscilloscope & X &  &  & \\ \hline
        Pressure platform &  & X &  & Accurate measurements would require coverage over a substantial area. \\ \hline
        Thermosensitive product & X &  &  & Not very precise\\ \hline
        Survey & X &  &  & \\ \hline
        Ruler/caliper & X &  &  & \\ \hline
        Thermocouple &  & X &  & Can be done if pre-welded \\ \hline
        Contact thermometer &  & X &  & Does not allow real-time study \\ \hline
        Ultrasonics &  & X &  & \\ \hline
        Video & X &  &  & Caution about what is intended to be measured \\ \hline
        Welding speed & X &  &  & \\ \hline
        VO2 &  &  & X & Little interest \\
       \hline
    \end{tabular}
    \caption{Summary of the measurement tools applicability in our study}
    \label{tab:measurement}
    \vspace{-1 cm}
\end{table}

Within these parameters, we can identify which one matches with a cognitive aspect of skills, meaning one pillar of the scheme. Consider, for instance, the electric parameters that an operator can set, embodying a prescribed rule of action. By examining the correlation between a torch manipulation reaction and the weld pool dimension, we identify an inference mechanism at play. Time reaction being unknown, we have to be careful when we analyze data. Through collaborative discussions with the operator and meticulous comparisons with other parameters, we can find and derive goals and operational invariants. This intricate analysis provides valuable insights into the interplay of cognitive skills and their manifestation within the defined framework.
\section{Survey and interview: A subjective measure of the rationale of the operator}
To gather the operator's insights, especially to understand the rationale behind their actions, we can utilize a survey or conduct an interview. However, a drawback of using a survey is its restrictive nature, providing limited freedom for the operator's responses and potentially directing them in a particular direction. This limitation could subsequently constrain the depth of the analysis we can perform.

Several interview techniques are available. Our objective is to comprehend the reasons behind specific actions, and for this purpose, Pierre Vermersch's explicitation interview appears most suitable \cite{vermersch_entretien_2019}. However, this method is acknowledged for its complexity in implementation, requiring training in interview techniques. Another equally interesting and easier-to-implement method is autoconfrontation.

Autoconfrontation, a commonly used technique, involves using traces, including videos, to place an operator back into the action situation, allowing them to explain their gestures \cite{theureau_les_2010}. Another method is the double instruction, aiming to prompt individuals to assume the role of an instructor to explain their profession \cite{laine_chapitre_2019}.

Interview techniques focus the interviewee's attention on specific aspects, such as action realization with video support in autoconfrontation, precise and non-generalized actions in the explicitation interview, and pedagogical debate in the case of double instruction. The importance of pauses remains central in these techniques. As this report concentrates on objective measurements, we won't delve further into these interview techniques.
\section{Proposed experimental protocol}
\begin{figure}
    \centering
    \includegraphics[width=0.5\linewidth]{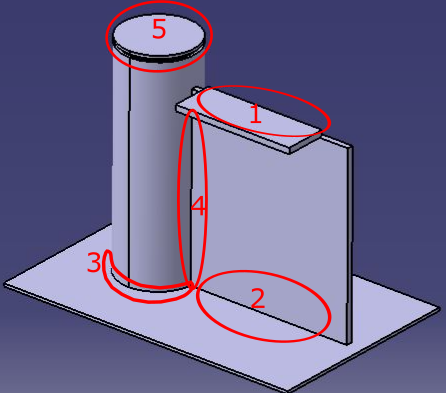}
    \caption{Reference piece to be used in the experimental protocol}
    \label{fig:Pièce référence}
\end{figure}

We propose creating a reference piece for a group of welders. This piece should encompass various situations, easily adaptable into a few variants (thickness and material), and representative of many mechanized welding production processes. To achieve this, we present the piece in Figure~\ref{fig:Pièce référence}.

This part incorporates the following situations (positions according to ISO 6947 \cite{iso_6947_soudage_2019}:
\begin{enumerate}
\item Angle at the ceiling – plate against plate (position PD)
\item Flat angle – plate against plate (position PB)
\item Flat angle – tube against plate (position PB)
\item Vertical ascending angle – tube against plate (position PF)
\item Flat angle – tube against tube (position PB - external)
\end{enumerate}

The welder will have at his disposal all the necessary sheets and tubes accompanied by the technical drawing of this reference part.  We propose to manufacture this piece with three different thicknesses: 2 mm, 4 mm, and 6 mm. A second piece may be produced in a subsequent trial to analyze edge-to-edge welds, especially in GTAW welding, to observe a task that is challenging for automation.

Before carrying out the welds, we will equip the welders with a set of inertial sensors to obtain the orientation of body segments (arms, forearms, torso, etc.). These inertial sensors can be incorporated into a suit (Teslasuit© or Movella©, or using IMUs with custom coding). We will focus on joint amplitudes, particularly those of the upper limbs, directing our attention to torch manipulation speed, pause times, and back-and-forth movements.

We will add a set of sensors to the work environment:
(i) Non-contact thermometer (infrared) to measure temperature,
(ii) A microphone to analyze sound,
(iii) A welding camera (Cavitar\textsuperscript{\textcopyright}) to analyze the fusion pool with filler metal and stick-out,
(iv) An RGB camera to qualitatively analyze the complete body movement and provide a record of the activity,
(v) Inertial measurement units to capture movements.

After completing the weld, we suggest conducting an exchange with the welder in the form of an interview to analyze the reasons behind their actions and understand their activity.

This article underscores the challenges and opportunities related to assessing welders' schemes in a welding environment. The interference of the electric arc (light, electromagnetic field, etc.) on most measurement tools justifies comparing objective indicators, such as sound and temperature, alongside subjective evaluations through self-confrontation interviews \cite{theureau_les_2010}. Our proposed approach involves analyzing welders' movements and reactions using objective measures like sound and temperature in various representational situations provided in a reference workpiece (see Figure \ref{fig:Pièce référence}). Our analysis will focus on identifying the control loop made by welders to operate a regulation during the welding process. The goal is to identify invariant organizational aspects across multiple situations, reflecting their schemes. By examining welders' movements and reactions with objective data, we aim to gain a comprehensive understanding of their schemes and develop a model encompassing all facets of their expertise. This model aspires to address current technical limitations faced by robots in specific tasks by incorporating human gestures and cognitive aspects of skills.


\section{Conclusions and future work}
The workforce in industries, including welders, constitutes the core of Industry 5.0, yet understanding the principles behind their skills poses challenges. In this article, we propose a set of measures to explore the reasoning behind each decision made by welders. Acknowledging the numerous variables influencing the welding process, we aim to identify critical ones used by welders, recognizing the presence of implicit skills, described within schemes framework.

Measuring a welder's movement is crucial but is a time-consuming step. Inertial Measurement Units (IMUs) offer a suitable compromise, despite requiring signal processing. While the weld pool dimensions are critical, measuring them poses challenges.

This article primarily centers on welding, yet the methodology employed here can be applicable to various professions. In summary, individuals seeking to analyze schemes must comprehend the intricacies of the job to determine multiple representative situations. Within these scenarios, it is imperative to select variables relevant to the operator. Then we have to measure them. Throughout the data analysis process, emphasis should be placed on identifying control loops that connect perception and action. Upon identification of these loops, it is crucial to verify their consistency across different situations. The organizational constants revealed through this process are indicative of the skills.
\bibliographystyle{splncs04}
\bibliography{bib-list}
\end{document}